\documentclass[conference]{IEEEtran}
\IEEEoverridecommandlockouts
\usepackage{cite}
\usepackage{amsmath,amssymb,amsfonts}
\usepackage{graphicx}
\usepackage{textcomp}
\usepackage{xcolor}
\def\BibTeX{{\rm B\kern-.05em{\sc i\kern-.025em b}\kern-.08em
    T\kern-.1667em\lower.7ex\hbox{E}\kern-.125emX}}
    
\usepackage{algorithm}
\usepackage{algpseudocode}
\usepackage{comment}
\usepackage{multirow}
\usepackage{booktabs}
\usepackage{longtable}

\begin{document}

\title{A Late Collaborative Perception Framework for 3D Multi-Object and Multi-Source Association and Fusion}

\author{
\IEEEauthorblockN{Maryem Fadili\textsuperscript{1,}\textsuperscript{2}\textsuperscript{*},
Mohamed Anis Ghaoui\textsuperscript{1},
Louis Lecrosnier\textsuperscript{2},
Steve Pechberti\textsuperscript{1}, 
Redouane Khemmar\textsuperscript{2}}

\IEEEauthorblockA{\textsuperscript{1}VEDECOM, Versailles, France}

\IEEEauthorblockA{\textsuperscript{2}IRSEEM, Saint-Etienne du Rouvray, France}

\textsuperscript{*}Corresponding author : maryem.fadili@vedecom.fr
}

\maketitle

\begin{abstract}
In autonomous driving, recent research has increasingly focused on collaborative perception based on deep learning to overcome the limitations of individual perception systems. Although these methods achieve high accuracy, they rely on high communication bandwidth and require unrestricted access to each agent’s object detection model architecture and parameters. These constraints pose challenges real-world autonomous driving scenarios, where communication limitations and the need to safeguard proprietary models hinder practical implementation.

To address this issue, we introduce a novel late collaborative framework for 3D multi-source and multi-object fusion, which operates solely on shared 3D bounding box attributes—category, size, position, and orientation—without necessitating direct access to detection models.

Our framework establishes a new state-of-the-art in late fusion, achieving up to five times lower position error compared to existing methods. Additionally, it reduces scale error by a factor of 7.5 and orientation error by half, all while maintaining perfect 100\% precision and recall when fusing detections from heterogeneous perception systems. These results highlight the effectiveness of our approach in addressing real-world collaborative perception challenges, setting a new benchmark for efficient and scalable multi-agent fusion.
\end{abstract}


\section{Introduction}
\label{sec:introduction}

Autonomous and automated driving is set to revolutionize transportation, making travel safer and more efficient. At the heart of this technology is a vehicle’s ability to accurately perceive and interpret its surroundings. To do this, autonomous systems rely on multiple sensors such as cameras and LiDARs, to detect obstacles and anticipate their movements. However, these technologies still face significant challenges, such as occlusions-particularly in urban and semi-urban environments-and adverse weather conditions can obstruct sensor performance, making it difficult for vehicles to consistently perceive their surroundings with full accuracy~\cite{caillot_survey_2022, hu_collaborative_2024}.

To overcome these limitations, collaborative perception is a solution to improve autonomous driving by enabling vehicles and the infrastructure to share data. Through vehicle-to-everything (V2X) communication, road agents exchange information to build a more complete view of their surroundings. This improves detection accuracy, extends perception range, and enhances safety. As a result, autonomous systems can better navigate complex and dynamic environments.

However, while most research efforts in collaborative perception adopt early or deep fusion strategies~\cite{zhao_coopre_2024, su_collaborative_2024}, these approaches often require high communication bandwidth or access to the parameters of the proprietary deep learning model, limiting practical implementation in real-world scenarios. Late fusion methods, by sharing only detected objects or object-level scene representations, drastically reduce communication overhead~\cite{abdali2025}. However, existing late fusion approaches typically rely on simplistic geometric rules for merging bounding boxes~\cite{neubeck_efficient_2006, zimmer_infradet3d_2023} and often struggle to maintain high accuracy under noisy detections or incomplete overlap. 
Furthermore, these methods are predominantly designed for 2D bounding box fusion \cite{fadili2025}, whereas full 3D bounding box information is essential to achieve robust collaborative perception in autonomous driving.

To address these limitations, we propose in this paper :
\begin{enumerate}
    \item A late collaborative framework for 3D bounding boxes association and fusion that leverages a novel Combined Score-Based Association (CSBA-3D) algorithm, enabling robust matching of objects across multiple agents despite uncertainties in position, orientation, and dimensions.
    \item A Weighted Least-Squares (WLS-3D) fusion method adapted to 3D bounding boxes, incorporating noise models for position, orientation, and object dimensions.
    \item A thorough evaluation on a \emph{pseudo-collaborative} dataset derived from nuScenes~\cite{caesar_nuscenes_2020,fadili2025}, where we inject controllable noise to emulate the varied detection quality of different agents.
\end{enumerate}

This paper is organized as follows: Section 2 reviews the related work, followed by Section 3, which details the methodology behind our association and fusion framework. Section 4 outlines the experimental setup, while Section 5 presents the results along with their analysis. Finally, Section 6 concludes the paper and discusses future research directions.

\section{Related Work}
\label{sec:related-work}
\textbf{Collaborative perception.}
Methods are divided into three general approaches: early, intermediate, and late fusion~\cite{caillot_survey_2022,han_collaborative_2023}. Early fusion methods~\cite{zhao_coopre_2024, chen_co3_2022} share raw sensor data (e.g., point clouds or images), thus requiring high-bandwidth communications and accurate sensor spatial alignment. 
Deep fusion~\cite{su_collaborative_2024,li_v2x-dgw_2024,qiao_cobevfusion_2023} among many others, exchanges intermediate features extracted from deep networks, with significant bandwidth usage and an exposure of intellectual proprietary or heterogeneous agent model parameters. In contrast, late fusion shares only final detection outputs or object-level data~\cite{DMSTrack,picard_decentralized, fadili2025}, making it more communication-efficient and preserving intellectual property rights which is compatible for integration across heterogeneous sensors from multiple vendors.

While late collaboration has traditionally shown lower accuracy than deep fusion methods due to its reliance on geometric strategies~\cite{han_collaborative_2023}, recent findings~\cite{abdali2025} demonstrate that late fusion can surpass intermediate fusion in both bandwidth efficiency and scalability—requiring only $2^{9}$ bytes per frame compared to $2^{19}$ bytes for deep fusion. 

Most late fusion approaches in autonomous driving and robotics employ Non-Maximum Suppression (\textit{NMS}) techniques to eliminate redundant detections. Standard NMS~\cite{neubeck_efficient_2006} retains only the highest-scoring detections based on Intersection over Union (\textit{IoU}) thresholds, sometimes suppressing valid detections. To mitigate this issue, Soft-NMS~\cite{bodla_soft-nms_2017,xu_model-agnostic_2023} reduces the scores of overlapping boxes instead of discarding them, while Weighted-NMS~\cite{shen_competitive_2021} merges detections using confidence-weighted averages. Other variations, such as Adaptive-NMS~\cite{liu_adaptive_2019}, which adjusts IoU thresholds based on object density, and Learning-NMS~\cite{hosang_learning_2017}, which leverages neural networks to optimize suppression patterns, still fundamentally rely on IoU. The 3D IoU variant is also used in state of the art late tracking method like \textit{DMSTrack} \cite{DMSTrack}, \textit{ADB3DMOT} \cite{weng_ab3dmot_2020} and \textit{V2V4Real} \cite{yang2023v2v4real} to associate multi-agent tracked objects. However, IoU-based methods struggle with small objects like pedestrians and penalize non-overlapping detections. To address these limitations, the Generalized IoU (\textit{GIoU}) metric ~\cite{giou2019} was introduced, extending IoU by incorporating the spatial proximity of non-overlapping bounding boxes through the smallest enclosing box.

Euclidean distance-based association is commonly used in late collaboration methods such as \textit{InfraDet3D-Late}~\cite{zimmer_infradet3d_2023}, \textit{DAIR-V2X-Late}~\cite{yu2022dairv2x}, and \cite{picard_decentralized}. \textit{InfraDet3D-Late} retains only the bounding box closest to the sensor, whereas \textit{DAIR-V2X-Late} and \cite{picard_decentralized} compute the fused bounding box as the average of associated detections.

Other association techniques, presented in the survey \cite{rakai2021survey}, include probabilistic methods such as Joint Probabilistic Data Association (JPDA) \cite{fortmann1983sonar} and Multiple Hypothesis Tracking (MHT) \cite{blackman2004multiple}, which utilize likelihood models to effectively track objects in crowded environments and handle occlusions. However, these methods are computationally intensive, making them impractical for real-time applications like autonomous driving. To address this, hybrid approaches \cite{zhang2008global, leal2016learning} integrate local detection-based association with global optimization techniques, enhancing accuracy and robustness in complex scenarios. Nevertheless, these methods often introduce significant computational overhead and require careful parameter tuning to achieve generalization.

To ensure a real-time application while handling uncertainties of agent detections, authors of \cite{fadili2025} proposed \textit{CSBA}, an association method that formulates a cost function by combining spatial proximity, size overlap, and yaw consistency. Additionally, they introduced a fusion framework, \textit{WLS}, which leverages weighted least squares followed by Kalman filter-based tracking. This approach, however limited to 2D bounding boxes projected onto the Bird’s Eye View plane, demonstrates robust performance in collaborative perception across varying detector qualities, outperforming state-of-the-art NMS-based fusion methods. 

Other \textbf{synchronous} data fusion techniques, beyond WLS, include Maximum Likelihood (ML) estimation~\cite{kay1993statistical}, which optimizes data likelihood under Gaussian noise but requires prior knowledge of measurement uncertainties. However, uncertainty estimation remains an underexplored area in object detection research~\cite{xu_model-agnostic_2023, su2023uncertainty, mun_uncertainty_2023}. Most state-of-the-art object detectors, which heavily rely on deep networks~\cite{li_yolov6_2023, focalformet3d, li_bevformer_2022}, typically output only bounding box attributes with confidence scores, without explicitly providing uncertainty estimates for position and orientation.\newline
\indent \textbf{Collaborative perception datasets.} Collaborative perception datasets leverage multiple viewpoints to enhance environmental understanding. Existing datasets fall into two main categories: simulated and real-world. Simulated datasets such as \textit{OPV2V} \cite{xu2022opv2v}, \textit{V2X-Sim}~\cite{li2021v2xsim}, and \textit{V2XSet}~\cite{hu2022v2xset} offer large-scale multi-agent scenarios but rely on artificial object dynamics, which may not fully capture real-world behavior. Real-world datasets like \textit{V2X-Seq}\cite{wang2023v2xseq} and \textit{V2V4Real}\cite{yang2023v2v4real} enable cooperative perception evaluation in realistic driving conditions. However, they often focus on limited perspectives, primarily V2I or V2V interactions, restricting generalization to broader multi-agent settings. Additionally, datasets such as \textit{V2V4Real} emphasize vehicle detection while overlooking vulnerable road users (VRUs), whereas \textit{TUMTraf}\cite{zimmer_tumtraf_2024} captures diverse objects in high-speed highway and intersection scenarios but lacks representation of low-speed, constant-velocity objects essential for urban driving.\newline
\indent Considering those limitations, we sought a dataset that provides realistic object dynamics, diverse object categories, and varied motion characteristics. While \textit{DAIR-V2X} initially appeared as a promising candidate, accessibility issues prevented its use (eg. broken links, inaccessibilty outside of china). The authors of \cite{fadili2025} propose a method to answer those limitations by constructing a \textit{pseudo-collaborative} dataset augmented from autonomous driving dataset like \textit{nuScenes} \cite{caesar_nuscenes_2020}. By introducing noises to annotated ground truths, we can simulate multi-agents perception data. This approach allows for precise control over noise while maintaining realistic object dynamics and categories, ensuring a fair and reproducible benchmark for late fusion evaluation. \newline
\indent In this work, we extend Weighted Least Squares to 3D bounding boxes and introduce an enhanced association method that moves beyond reliance on 2D object matching. We focus on evaluating our association-fusion pipeline in the validation set of \textit{pseudo-collaborative} nuScenes dataset as presented in \cite{fadili2025}. Our framework being a late collaborative method, comparison with deep fusion methods like \cite{qiao_cobevfusion_2023, hu_where2comm_2022}... is out of scope because we assume we have no access to raw and intermediate features. 

\section{3D Multi-Object and Multi-Source Association and Fusion} 
\label{sec:method}

We present a novel late fusion pipeline for 3D object association and fusion. 

Following previous works on uncertainty estimation for object detectors~\cite{mun_uncertainty_2023, LeUncertainty2018, su2023uncertainty}, we assume that all noise sources follow a Gaussian distribution.

Furthermore, we assume that each agent exclusively provides object-level information in the form of 3D bounding boxes, including position, size, orientation, and category. Additionally, we consider \textbf{object classifications to be consistent} across agents, ensuring there is no classification ambiguity or confusion in the fusion process. 

To account for asynchronous agents and communication-induced latencies, we employ a sliding time window that aggregates detections within a predefined temporal interval $\Delta t$. This approach aligns data from multiple sources—despite slight timing offsets—enabling reliable fusion while maintaining temporal consistency.

In this pipeline, we first match bounding boxes from different agents using our novel Combined Score-Based Association (CSBA-3D) and then merge them using Weighted Least Squares fusion algorithm (WLS-3D). 

\subsection{State representation of 3D bounding boxes}
\label{ssec:state_rep}
Each detected 3D bounding box $i$ from agent $n \in \{1,\dots,N\}$ at time $k$ is given by:
\begin{equation}
    \mathbf{y}_k^{(n,i)} = \big(x_k,\, y_k,\, z_k,\, l_k,\, w_k,\, h_k,\, \theta_k\big) + \mathbf{e}_k^{(n,i)},
\end{equation}
where $(x_k, y_k, z_k)$ is the center position of the bounding box in a global or local coordinate system (e.g., aligned with the ego-vehicle or a map frame), $l_k,\, w_k,\, h_k$ represent the length, width, and height of the box, and $\theta_k$ is the yaw angle.%

The noise term $\mathbf{e}_k^{(n,i)}$ is assumed to follow a Gaussian distribution with zero mean and covariance $\mathbf{R}_k^{(n,i)}$.

We aim to fuse $M$ detections referring to the same object, resulting in a single fused state:
\begin{equation}
    \mathbf{z}_k = \big(\bar{x}_k,\, \bar{y}_k,\, \bar{z}_k,\,
    \bar{l}_k,\, \bar{w}_k,\, \bar{h}_k,\, \bar{\theta}_k\big).
\end{equation}
Below, we describe how to \textbf{associate} bounding boxes from multiple agents and \textbf{combine} them into one fused detection.

\subsection{3D Combined Score-Based Association (CSBA-3D)}
\label{ssec:CSBA-3D}
To match bounding boxes from different agents at a given time step $k$, we introduce Combined Score-Based Association (CSBA-3D) method that extends CSBA~\cite{fadili2025} to handle 3D Objects. The algorithm is detailed in Alg.\ref{alg:CSBA-3D}. 
Consider two sets of detections $\{\mathbf{y}_k^{(1,i)}\}_{i=1}^{I}$ and $\{\mathbf{y}_k^{(2,j)}\}_{j=1}^{J}$ from different sources. Our goal is to associate detections corresponding to the same physical object by solving a global assignment problem, formulated as:
\begin{equation}
    \underset{\mathbf{X}}{argmin} \quad \sum_{i=1}^{I} \sum_{j=1}^{J} X_{i,j} C(\mathbf{y}_k^{(1,i)}, \mathbf{y}_k^{(2,j)})
\end{equation}
where $X_{i,j} \in \{0,1\}$ is a binary variable indicating whether detection $i$ from source 1 is matched to detection $j$ from source 2, and $C(\mathbf{y}_k^{(1,i)}, \mathbf{y}_k^{(2,j)})$ is the pairwise cost. The optimization is solved using the Jonker-Volgenant assignment algorithm~\cite{jonker_shortest_1987}, which efficiently finds the minimum-cost matching.

\textbf{Score Definitions.} 
We introduce three scores to compare pairs of 3D bounding boxes:
\begin{itemize}
    \item \textit{Dimension Score (DS)} a novel score that quantifies the similarity between object dimensions. It extends \cite{hu_joint_nodate, fadili2025} to account for volumes while incorporating uncertainties.
    \item \textit{Center Score (CS)} a novel score that measures the positional alignment of bounding boxes using the Mahalanobis distance. It also refines \textit{CS} from \cite{hu_joint_nodate, fadili2025} to account for 3D distances while incorporating uncertainties.
    \item \textit{Orientation Score (OS)} introduced in \cite{fadili2025}, evaluates the angular similarity of object yaw orientations. 
\end{itemize}
Each score is normalized to lie in $[0,\,1]$, where $1$ represents a perfect match.

\noindent
{\bf Dimension Score (DS).} 
We define the volume of predicted and ground truth bounding boxes $V^{(1,i)}$ and $V^{(2,j)}$ as:

\begin{equation}
    V^{(1,i)} = l^{(1,i)} w^{(1,i)} h^{(1,i)}
\end{equation}
Next, we define the ratio and inverse ratio between both volumes, along with their uncertainties using first-order uncertainty propagation \cite{taylor1980}:

\begin{equation}
    r = V^{(1,i)}/V^{(2,j)}, \quad
    \sigma_r = r \sqrt{\left(\frac{\sigma_{V^{(1,i)}}}{V^{(1,i)}}\right)^2 + \left(\frac{\sigma_{V^{(2,j)}}}{V^{(2,j)}}\right)^2}
\end{equation}
where the volume uncertainties \( \sigma_{V^{(1,i)}} \) and \( \sigma_{V^{(2,j)}} \) are obtained also, using uncertainty propagation formula:
\begin{equation}
    \sigma_{V^{(1,i)}} = V^{(1,i)} \cdot \sqrt{\left(\frac{\sigma_{l^{(1,i)}}}{l^{(1,i)}}\right)^2 + \left(\frac{\sigma_{w^{(1,i)}}}{w^{(1,i)}}\right)^2 + \left(\frac{\sigma_{h^{(1,i)}}}{h^{(1,i)}}\right)^2}
\end{equation}
The Dimension Score (DS) is then computed using a Gaussian transformation to the minimum of Z-scores squares:
\begin{equation}
    \text{DS} = \exp \left( -\frac{\min(Z_r^2, Z_{r^{-1}}^2)}{2} \right)
\end{equation}
where Z-scores are computed as :
\begin{equation}
    Z_r = \frac{r - 1}{\sigma_r}, \quad
    Z_{r^{-1}} = \frac{r^{-1} - 1}{\sigma_{r^{-1}}}
\end{equation}
Uncertainty of inverse ratio is : $ \sigma_{r^{-1}} = {\sigma_r}$. Here, \( \sigma_r \) ensures that uncertainties in the volume ratio are properly incorporated into the score computation. A higher \( \sigma_r \) results in lower confidence in the dimension match, thereby reducing the \textit{Dimension Score}.

\noindent
{\bf Center Score (CS).} 
To evaluate positional consistency, we use the Mahalanobis distance:
\begin{equation}
    d_M = \sqrt{(\mathbf{p}^{(1,i)} - \mathbf{p}^{(2,j)})^T \mathbf{\Sigma}^{-1} (\mathbf{p}^{(1,i)} - \mathbf{p}^{(2,j)})}
\end{equation}
where $\mathbf{p} = [x, y, z]^T$ represents object center positions, and $\mathbf{\Sigma}$ is the combined covariance matrix of both detections. The final CS score is:
\begin{equation}
    \text{CS} = 1 - \frac{d_M}{\lambda_{\text{max}}}
\end{equation}
where $\lambda_{\text{max}}$ is a user-defined maximum distance threshold in meters.

\noindent
{\bf Orientation Score (OS).} 
Yaw similarity is computed based on the cosine similarity between orientation vectors and incorporating uncertainties as formulated in \cite{fadili2025} :
\begin{equation}
    \alpha_{i} = \frac{\theta^{(1,i)}}{\sigma(\theta^{(1,i)})}, \quad \alpha_{j} = \frac{\theta^{(2,j)}}{\sigma(\theta^{(2,j)})}
\end{equation}

where \( \theta^{(1,i)} \) and \( \theta^{(2,j)}\) are the respective yaw angles of the two observations in \textit{radians}, and \( \sigma(\theta^{(1,i)}) \) and \(  \sigma(\theta^{(2,j)}) \) are their associated uncertainties.

The OS score is then computed using the cosine similarity between the adjusted angles:

\begin{equation}
    \text{OS} = \frac{1 + \cos(\alpha_{i} - \alpha_{j})}{2}
\end{equation}

\textbf{Final Association Cost.} 
The pairwise cost function combines DS, CS, and OS using weighted summation:
\begin{equation}
\scriptsize
    C(\mathbf{y}_k^{(1,i)}, \mathbf{y}_k^{(2,j)} )= \frac{w_{ds}(1-\mathrm{DS}) + w_{cs}(1-\mathrm{CS}) + w_{os}(1-\mathrm{OS})}{w_{ds} + w_{cs} + w_{os}}
\end{equation}
where $w_{ds}, w_{cs}, w_{os}$ are user-defined weights. Higher scores correspond to lower costs, ensuring that well-matched objects are prioritized.

\textbf{The computational complexity} of the \textit{CSBA-3D} algorithm primarily depends on two stages: computing the pairwise cost matrix and solving the assignment problem. The cost matrix calculation has a complexity of $\mathcal{O}(|\mathcal{I}| \cdot |\mathcal{J}|)$, where $|\mathcal{I}|$ and $|\mathcal{J}|$ denote the cardinalities of the detection sets to be associated. Solving the assignment problem using classical methods such as the Hungarian algorithm has a worst-case complexity of $\mathcal{O}(\max(|\mathcal{I}|, |\mathcal{J}|)^3)$. However, by adopting the Jonker-Volgenant algorithm, the practical average-case complexity is significantly reduced, typically approaching $\mathcal{O}(\max(|\mathcal{I}|, |\mathcal{J}|)^2)$. Thus, the overall \textit{CSBA-3D} complexity is $\mathcal{O}(|\mathcal{I}| \cdot |\mathcal{J}| + \max(|\mathcal{I}|, |\mathcal{J}|)^2)$.

\textbf{Extending to $N>2$ agents.} 
For multi-agent fusion, detections can be sequentially associated in a pairwise manner across agents.

\makeatletter
\renewcommand{\ALG@beginalgorithmic}{\normalfont\ttfamily} 
\makeatother

\begin{algorithm}[H]
\footnotesize
\caption{CSBA-3D }
\label{alg:CSBA-3D}
\begin{algorithmic}[1]
\State \textbf{Input:} Two sets of detections $\mathcal{I}$, $\mathcal{J}$; Cost weights $(w_{ds}, w_{cs}, w_{os})$; Distance threshold $\lambda_{\text{max}}$
\State \textbf{Output:} Matched pairs $\mathcal{M}$, unmatched detections $\mathcal{U}_\mathcal{I}, \mathcal{U}_\mathcal{J}$
\State Initialize cost matrix $C \in \mathbb{R}^{|\mathcal{T}| \times |\mathcal{D}|}$
\ForAll{${\mathbf{y}_k^{(1,i)}} \in \mathcal{I}$}
    \ForAll{$\mathbf{y}_k^{(2,j)}\in \mathcal{J}$}
        \State Compute $\mathrm{DS} \leftarrow \text{DimensionScore}({\mathbf{y}_k^{(1,i)}}, \mathbf{y}_k^{(2,j)})$
        \State Compute $\mathrm{CS} \leftarrow \text{CenterScore}({\mathbf{y}_k^{(1,i)}}, \mathbf{y}_k^{(2,j)}, \lambda_{\text{max}})$
        \State Compute $\mathrm{OS} \leftarrow \text{OrientationScore}({\mathbf{y}_k^{(1,i)}}, \mathbf{y}_k^{(2,j)})$
        \State Compute cost:
        \State $C[i,j] \leftarrow \frac{w_{ds}(1-\mathrm{DS}) + w_{cs}(1-\mathrm{CS}) + w_{os}(1-\mathrm{OS})}{w_{ds}+w_{cs}+w_{os}}$
    \EndFor
\EndFor
\State Solve assignment: $(\mathcal{M}_i, \mathcal{M}_j) \leftarrow \text{LinearSumAssignment}(C)$
\State  $\mathcal{M} \leftarrow (\mathcal{M}_i, \mathcal{M}_j)$
\State $\mathcal{U}_\mathcal{I} \leftarrow \{\text{unmatched } \mathbf{y}_k^{(1,i)} \in \mathcal{I}\}$
\State $\mathcal{U}_\mathcal{J} \leftarrow \{\text{unmatched }\mathbf{y}_k^{(2,j)} \in \mathcal{J}\}$
\State \Return $\mathcal{M},\, \mathcal{U}_\mathcal{I},\, \mathcal{U}_\mathcal{J}$
\end{algorithmic}
\end{algorithm}

\subsection{Weighted Least-Squares for 3D bounding box fusion (WLS-3D)}
\label{ssec:wls}
Once bounding boxes across agents are associated, we merge them into a single fused detection. Suppose a matched group $\{\mathbf{y}_k^{(n,i_n)}\}_{n=1}^M$ that describes the same object. A Weighted Least-Squares (WLS) approach estimates the fused state $\hat{\mathbf{z}}_k$ by:
\begin{equation}
    \hat{\mathbf{z}}_k = 
    \biggl(\sum_{n=1}^{M}\mathbf{R}_k^{(n,i_n)\!-\!1}\biggr)^{-1} \sum_{n=1}^{M}\mathbf{R}_k^{(n,i_n)\!-\!1}\,\mathbf{y}_k^{(n,i_n)}
\end{equation}
where $\mathbf{R}_k^{(n,i_n)}$ is the covariance of detection $\mathbf{y}_k^{(n,i_n)}$. 
The covariance of the fused estimate is:
\begin{equation}
    \mathbf{P}_{\hat{\mathbf{z}}_k} 
    = \biggl(\sum_{n=1}^{M}\mathbf{R}_k^{(n,i_n)\!-\!1}\biggr)^{-1}
\end{equation}
In practical object detection pipelines, full covariance estimates for $(x,y,z,l,w,h,\theta)$ are typically unavailable. Instead, these uncertainties are either approximated or learned using methods such as detection scores or sensor models. While recent research has begun addressing this challenge, it remains an evolving area and has yet to reach full maturity (cf. \ref{sec:related-work}).

\section{Experiments} 
\label{sec:experiments}
We present quantitative evaluations of our proposed method with \textit{CSBA-3D} based association followed by Weighted Least-Squares 3D bounding box fusion \textit{WLS-3D}. We compare against state-of-the-art \textbf{late fusion} baselines.

It is important to note that our work focuses on late collaboration perception with minimal data exchange, relying exclusively on object-level 3D bounding boxes. Unlike feature-level fusion approaches, such as \cite{zimmer_tumtraf_2024} or \cite{su_collaborative_2024}, which require sharing raw sensor data or intermediate feature maps, our method operates under the assumption that such information is unavailable. Instead, we treat each agent’s detector as a black box, making no assumptions about its internal architecture. Therefore, the following experiments and results sections do not include comparisons with intermediate collaborative perception algorithms.

\subsection{Dataset}
We follow \cite{fadili2025} to build a \textbf{pseudo-collaborative} dataset using nuScenes ground truth annotations. We evaluate our approach on the complete nuScenes validation dataset, which includes $150$ scenes and approximately $145,000$ annotated objects. Our analysis covers all nuScenes object categories, such as cars, buses, trucks, pedestrians, bicycles, and motorcycles, among others.  We also store a synthetic \emph{confidence score} as in \cite{fadili2025} to mimic a detection confidence for each box, used by baseline NMS-based fusion methods for comparisons.

\subsection{Noise configuration} 
To ensure a comprehensive evaluation, we simulate bounding boxes detected by two agents positioned at a random distance apart. Following~\cite{fadili2025}, we introduce controlled noise by sampling errors from zero-mean Gaussian distributions to simulate a range of detector qualities. As detailed in Table~\ref{tab:noise_levels}, the noise configurations correspond to representative classes of perception systems with \textit{mild}, \textit{moderate}, and \textit{large} noise configurations. Specifically, standard deviations for positional errors are set to $0.5m$, $1.5m$, and $3m$, reflecting the increasing positional uncertainty from high-performance LiDAR-camera fusion systems to monocular setups and degraded sensors~\cite{Gahlert2020, lidar_camera_fusion, sensor_failures_survey}. Similarly, yaw errors are modeled with standard deviations of $5^\circ$, $20^\circ$, and $60^\circ$, while scale uncertainties (width, depth, height) follow standard deviations of $0.1$, $0.5$, and $1.0$, respectively. These noise levels directly mirror the characteristics of off-the-shelf, optimized, and degraded detectors, providing a realistic and systematic framework for evaluating fusion robustness under diverse uncertainty conditions.

We also apply a temporal sliding window of $\Delta t = 100\text{ms}$ to address asynchronous sensors and communication-induced latencies, knowing that the simulated dataset is annotated at a frequency of 2 Hz.

\begin{table}[h]
    \centering
    \caption{Noise configurations for position, yaw, and size perturbations applied to ground truth annotations from nuScenes.}
    \label{tab:noise_levels}
    \begin{tabular}{|l|c|c|c|}
        \hline
        \textbf{Noise Level} & \textbf{Std Position~(m)} & \textbf{Std Yaw~($^\circ$)} & \textbf{Std Scale} \\
        \hline
        Mild      & $0.5$  & $5^\circ$   & $0.1$  \\
        Moderate  & $1.5$  & $20^\circ$  & $0.5$  \\
        Large     & $3.0$  & $60^\circ$  & $1.0$  \\
        \hline
    \end{tabular}
\end{table}

\subsection{Metrics}
\label{ssec:metrics}
After bounding box fusion, predicted boxes are matched to ground truth using object identifiers. If multiple predictions align with the same ground truth, only the closest one is counted as a true positive, while the others are classified as false positives.

We evaluate our approach using metrics adapted from the nuScenes benchmark~\cite{caesar_nuscenes_2020}, including classical Precision and Recall to assess detection performance. The Mean Average Translation Error (mATE) measures the localization performance and the Mean Average Orientation Error (mAOE) quantifies the orientation prediction errors. Lastly, we use Mean Average Scale Error (mASE) as introduced in \cite{fadili2025} evaluates the discrepancy in object dimensions by computing the Euclidean distance between the predicted and ground truth values of length, width, and height.

While the original nuScenes metrics focus purely on \emph{true positives}, we modify them to also penalize false positives, ensuring that approaches with fewer false positives do not appear worse by ignoring unmatched predictions.

\subsection{Baselines}
We implemented three state of the art fusion methods augmenting them to handle 3D objects : Standard NMS as in \cite{neubeck_efficient_2006}—used in approaches such as DMSTrack\cite{DMSTrack}, AB3DMOT~\cite{weng_ab3dmot_2020}, and V2V4Real~\cite{yang2023v2v4real}—, Promote-Suppress Aggregation (\textit{PSA})~\cite{xu_model-agnostic_2023}, Weighted Box Fusion (\textit{WBF})~\cite{shen_competitive_2021}, and GIoU-Based NMS from \cite{giou2019}. We also compare to \textit{InfraDet3D} \cite{zimmer_infradet3d_2023} and \textit{DAIR-V2X} ~\cite{yu2022dairv2x} \textbf{late fusion} algorithm using a threshold distance of \textbf{$3m$} as stated in both papers.
We also evaluate two variants of our pipeline :
\begin{itemize}
    \item \textbf{\textit{WLS-3D w/ CSBA-3D}} : \textit{CSBA-3D} based association followed by Weighted-Least-Squares 3D \textit{WLS-3D} fusion;
    \item \textbf{\textit{WLS-3D w/ GT-Assoc}} : Perfect association using Ground Truth followed by followed by Weighted-Least-Squares 3D \textit{WLS-3D} fusion.
\end{itemize}
\textbf{CSBA parameters.} Weights $w_{ds}$, $w_{cs}$, and $w_{os}$ are set, respectively to $0.2$, $0.5$ and $0.3$, giving greater importance to spatial proximity. For fusion association, the scale $\mathbf{\lambda_{\text{max}}}$ applied to \textit{Center Score (CS)} is set to $6\times Std Position$ from Table \ref{tab:noise_levels}. 

\section{Results and Discussion}
\label{sec:results}
\subsection{Quantitative Results}
\begin{table*}[t]
\centering
\caption{Comparison of \emph{late fusion} methods under three different noise levels. Our WLS-3D consistently reduces errors (mATE, mAOE, mASE) while improving precision and recall compared to baseline methods.}
\label{tab:performance_metrics_all_dataset}
\begin{tabular}{llccccc}
\toprule
\textbf{Noise Level} & \textbf{Method} & \textbf{mATE (m)} & \textbf{mASE (m)} & \textbf{mAOE (deg)} & \textbf{Precision} & \textbf{Recall} \\
\midrule
\multirow{9}{*}{\textbf{Mild Noise}} 
    & NMS-STD-3D \cite{neubeck_efficient_2006} & 2.62 & 0.82 & 12.91 & 0.53 & 0.99 \\ 
    & PSA \cite{xu_model-agnostic_2023} & 2.74 & 2.31 & 13.66 & 0.50 & \textbf{1.00} \\
    & WBF \cite{shen_competitive_2021} & 2.60 & 0.80 & 12.86 & 0.53 & 0.99 \\
    & NMS-GIoU-3D \cite{giou2019} & 2.67 & 0.85 & 13.19 & 0.52 & 0.9 \\
    & InfraDet3D-Late~\cite{zimmer_infradet3d_2023} & 1.65 & 0.43 & 8.68 & 0.88 & \textbf{1.00} \\
    & DAIR-V2X-Late~\cite{yu2022dairv2x} & 1.42 & 0.42 & 7.19 & 0.88 & \textbf{1.00} \\
    & WLS-3D w/ CSBA-3D (Ours) & \textbf{0.99} & \textbf{0.34} & \textbf{5.61} & \textbf{1.00} & \textbf{1.00} \\
    \cmidrule(lr){2-7} 
    & WLS-3D w/ GT-Assoc (Ours) & 0.97 & 0.34 & 4.80 & 1.00 & 1.00 \\
\midrule
\multirow{9}{*}{\textbf{Moderate Noise}} 
    & NMS-STD-3D \cite{neubeck_efficient_2006} & 5.95 & 3.71 & 37.52 & 0.50 & \textbf{1.00} \\
    & PSA \cite{xu_model-agnostic_2023} & 5.94 & 3.73 & 37.51 & 0.50 & \textbf{1.00} \\
    & WBF \cite{shen_competitive_2021} & 5.94 & 3.71 & 37.39 & 0.51 & 0.99 \\
    & NMS-GIoU-3D \cite{giou2019} & 5.95 & 3.72 & 37.46 & 0.50 & \textbf{1.00} \\
    & InfraDet3D-Late~\cite{zimmer_infradet3d_2023} & 5.14 & 3.00 & 32.10 & 0.61 & \textbf{1.00} \\
    & DAIR-V2X-Late~\cite{yu2022dairv2x} & 5.06 & 3.01 & 30.42 & 0.61 & \textbf{1.00} \\
    & WLS-3D w/ CSBA-3D (Ours) & \textbf{2.34} & \textbf{1.36} & \textbf{16.66} & \textbf{1.00} & \textbf{1.00} \\
    \cmidrule(lr){2-7} 
    & WLS-3D w/ GT-Assoc (Ours) & 2.10 & 1.35 & 13.24 & 1.00 & 1.00 \\
\midrule
\multirow{9}{*}{\textbf{Large Noise}} 
    & NMS-STD-3D \cite{neubeck_efficient_2006} & 10.70 & 6.35 & 100.45 & 0.51 & 0.99 \\
    & PSA \cite{xu_model-agnostic_2023} & 10.75 & 6.42 & 100.93 & 0.50 & \textbf{1.00} \\
    & WBF \cite{shen_competitive_2021} & 10.69 & 6.34 & 100.18 & 0.50 & 0.99 \\
    & NMS-GIoU-3D \cite{giou2019} & 10.74 & 6.40 & 100.89 & 0.50 & 0.99 \\
    & InfraDet3D-Late~\cite{zimmer_infradet3d_2023} & 10.31 & 6.06 & 95.76 & 0.53 & \textbf{1.00} \\
    & DAIR-V2X-Late~\cite{yu2022dairv2x} & 10.29 & 6.03 & 95.25 & 0.53 & \textbf{1.00} \\
    & WLS-3D w/ CSBA-3D (Ours) & \textbf{4.83} & \textbf{2.35} & \textbf{42.47} & \textbf{1.00} & \textbf{0.98} \\
    \cmidrule(lr){2-7} 
    & WLS-3D w/ GT-Assoc (Ours) & 3.81 & 2.35 & 35.76 & 1.00 & 1.00 \\
\midrule
\multirow{9}{*}{\textbf{Mild Noise + Large Noise}} 
    & NMS-STD-3D \cite{neubeck_efficient_2006} & 6.74 & 3.66 & 57.23 & 0.50 & 0.99 \\
    & PSA \cite{xu_model-agnostic_2023} & 6.74 & 3.65 & 57.22 & 0.50 & \textbf{1.00} \\
    & WBF \cite{shen_competitive_2021} & 6.75 & 3.65 & 57.24 & 0.50 & 0.99 \\
    & NMS-GIoU-3D \cite{giou2019} & 6.73 & 3.65 & 57.24 & 0.50 & 0.99 \\
    & InfraDet3D-Late~\cite{zimmer_infradet3d_2023} & 6.28 & 3.30 & 51.85 & 0.53 & \textbf{1.00} \\
    & DAIR-V2X-Late~\cite{yu2022dairv2x} & 6.26 & 3.31 & 52.02 & 0.56 & \textbf{1.00} \\
    & WLS-3D w/ CSBA-3D (Ours) & \textbf{1.36} & \textbf{0.44} & \textbf{23.50} & \textbf{1.00} & \textbf{1.00} \\
    \cmidrule(lr){2-7} 
    & WLS-3D w/ GT-Assoc (Ours) & 1.32 & 0.44 & 17.02 & 1.00 & 1.00 \\
\bottomrule
\end{tabular}
\end{table*}

Table \ref{tab:performance_metrics_all_dataset} shows a representative comparison of the methods under the three noise levels: \textbf{mild noise}, \textbf{moderate noise}, and \textbf{large noise}.

Under mild noise conditions, even strong baselines such as \textit{InfraDet3D-Late} and \textit{DAIR-V2X-Late} achieve relatively low $mATE$ values of $1.65m$ and $1.42m$, respectively. However, \textit{WLS-3D w/ CSBA-3D} further reduces translation error to $0.99m$  while achieving a $5.61^\circ$ $mAOE$, outperforming all baselines. Notably, replacing \textit{CSBA-3D} with ground-truth association (\textit{WLS-3D w/ GT-Assoc}) results in only a slight orientation improvement to $4.80^\circ$, suggesting that \textit{CSBA-3D} already achieves near-optimal matching in this low-noise regime.  

With moderate noise, traditional \textit{3D NMS} approaches (e.g., \textit{NMS-STD-3D}, \textit{WBF}) yield significantly higher errors, with $mATE \approx 5.9$–$6.0m$  and $mAOE \approx 37^\circ$. \textit{InfraDet3D-Late} and \textit{DAIR-V2X-Late} struggle, producing $\sim 5m$ for $mATE$ and $3m$ error in size estimation. In contrast, \textit{WLS-3D w/ CSBA-3D} achieves a much lower $2.34m$  $mATE$ and $16.66^\circ$ $mAOE$, while maintaining perfect precision and recall ($1.00$). \textit{CSBA-3D} remains highly effective even in moderate noise scenarios with only $0.24m$(position), $0.01m$(size) and $3^\circ$(yaw) errors introduced due to association imperfections.

In the large-noise setting, where detections are highly unreliable, \textit{baseline fusion approaches} yield $mATE$ values exceeding $10m$ and $mAOE$ approaching $100^\circ$, with \textit{InfraDet3D-Late} and \textit{DAIR-V2X-Late} maintaining translation errors around $10m$. This significantly surpasses the expected bounds dictated by uncertainty propagation principles, where an algorithm’s errors should ideally remain within $3\sigma$—and preferably within $1\sigma$—of the estimated detection uncertainty for reliable validation. In stark contrast, \textit{WLS-3D w/ CSBA-3D} more than halves these errors, achieving $4.83m$ in $mATE$, $2.35m$ in $mASE$ and $42.47^\circ$ in $mAOE$, while still preserving high precision ($1.00$) and recall ($1.00$). Even when utilizing ground-truth association (\textit{WLS-3D w/ GT-Assoc}), improvements are incremental ($3.81m$ , $35.76^\circ$), underscoring \textit{CSBA-3D’s} robustness in handling significant detector noise and achieving results close to the theoretical best of \textit{WLS-3D}. It is noteworthy that \textit{CSBA-3D} does not introduce errors in the estimated bounding box dimensions, demonstrating the effectiveness of the \textit{Dimension Score} in preserving object size consistency.

Furthermore, $recall = 1.00$ implies zero false negatives, whereas $precision = 1.00$ indicates zero false positives. A closer inspection of \ref{tab:performance_metrics_all_dataset} reveals that \textit{WLS-3D w/ CSBA-3D} consistently achieves these perfect detection metric, which indicates that all the detections are efficiently associated and merged. In contrast, baselines such as \textit{NMS-STD-3D} and \textit{WBF} frequently report $precision \approx 0.50$–$0.53$, indicating suboptimal merging strategies that retain too many duplicate detections. This issue is particularly evident for small objects, such as pedestrians and bicycles, where IoU-based and distance-based (with fixed threshold) association strategy used in baselines, struggles to handle detection misalignment effectively, leading to inflated false positives.

Now, what happens when an object detected by an agent with high uncertainty is fused with an object detected by an agent with low uncertainty (e.g., a fusion of a LiDAR and a camera operating in low-light conditions) ? This is where our proposed \textit{WLS-3D w/ CSBA-3D} excels, achieving approximately five times lower $mATE$ compared to baseline methods. Additionally, the scale error is reduced by a factor of $7.5$, and the orientation error is reduced by half, all while maintaining a perfect precision and recall of $1$. These results demonstrate the robustness of our framework in handling heterogeneous sources by effectively leveraging their respective uncertainty estimates.

It is important to note that our fusion depends entirely on the detector's performance: it does not create or suppress information. If the fusion is not performed for a given set of objects (e.g., because two objects from different sources have not been associated by \textit{CSBA-3D}), both detections are preserved—even if this results in a false positive. In safety-critical applications like autonomous driving, we tend to prefer false positives over false negatives. 

\subsection{Qualitative Results}
\begin{figure}[h!]
  \centering
  \vbox{
    \centering
    \includegraphics[scale=0.3]{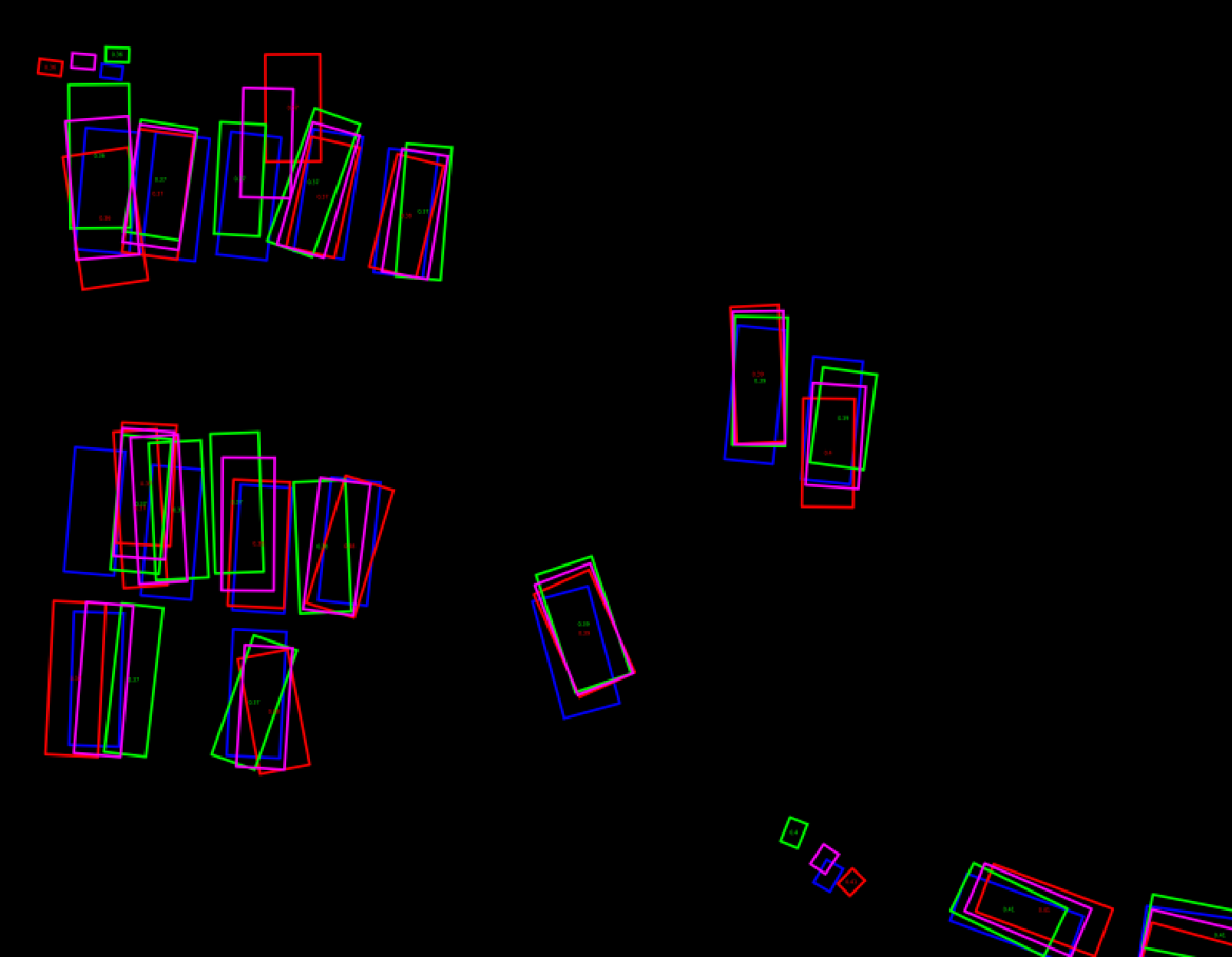}
  }  
  \vspace{0.3cm} 
  \vbox{
    \centering
    \includegraphics[scale=0.3]{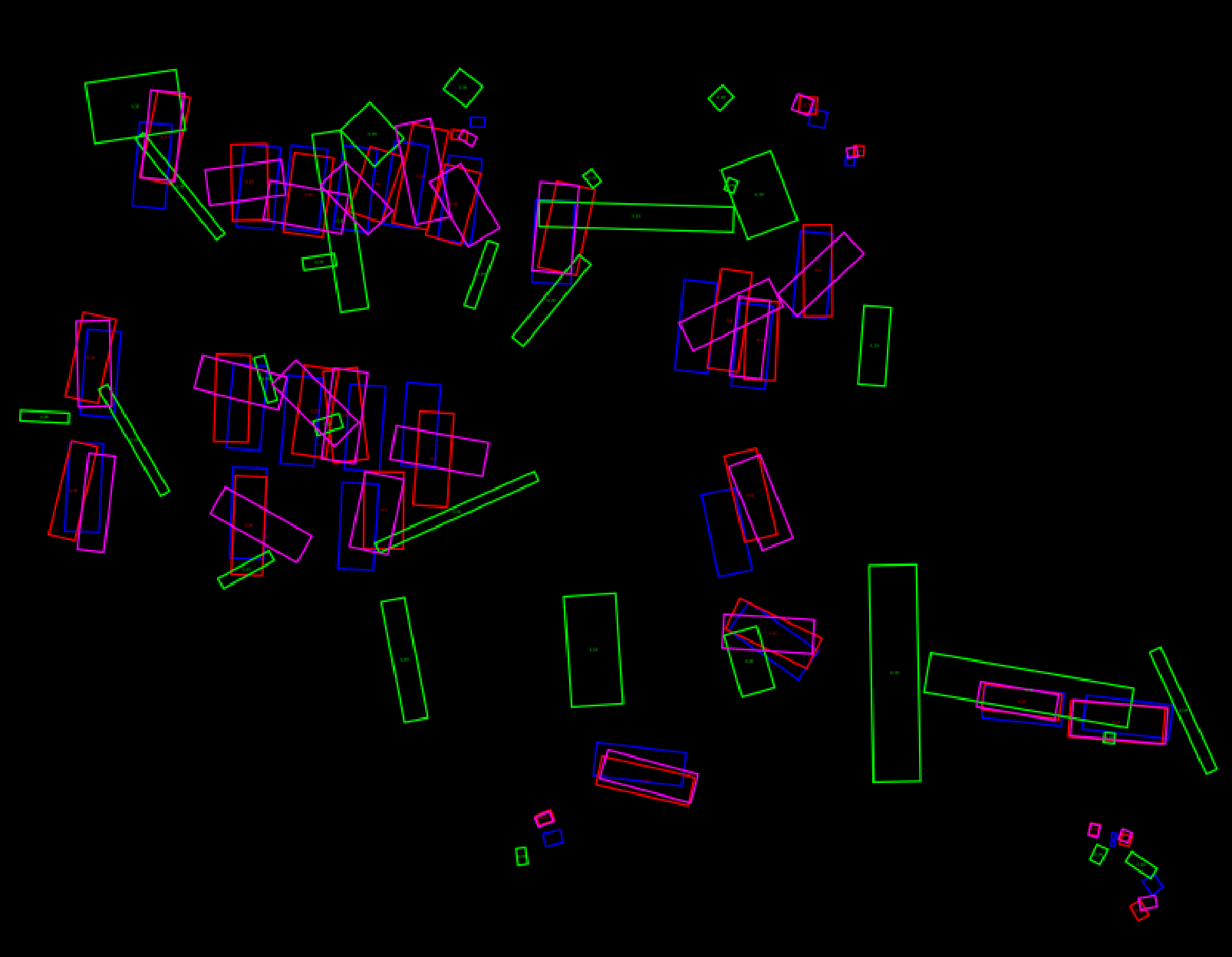}
  } 
  \caption{ Qualitative results of our association algorithm, \textit{CSBA-3D}, and fusion algorithm, \textit{WLS-3D}, in handling varying levels of noise in detections. The \textbf{top} figure shows results for the fusion of \textbf{mild noise} detections, while the \textbf{bottom} figure illustrates the fusion of both \textbf{high-noise and low-noise detections}. The visualizations include \textcolor{blue}{ground truth}, \textcolor{red}{Agent 1 detections}, \textcolor{green}{Agent 2 detections}, and \textcolor{magenta}{fused objects}. (Best viewed in color)}
  \label{fig:results}
\end{figure}

Figure \ref{fig:results} illustrates the performance of our association algorithm, \textit{CSBA-3D}, and fusion algorithm, \textit{WLS-3D}. The top image highlights the algorithm's effectiveness in a mild noise context, where the fused objects closely align with the ground truth, demonstrating the high accuracy of our approach.

Our framework perform well in handling a combination of high-noise and low-noise detections (\textbf{Bottom figure} of Figure \ref{fig:results}). The results indicate that the fused size and position are accurately estimated by our algorithm, particularly for large objects, while performance is slightly less effective for smaller objects. However, the results for orientation are less robust, especially in the presence of significant noise in the detections, with errors reaching up to $180^\circ$ in some cases. This limitation could be mitigated by incorporating a tracking algorithm, as demonstrated by \cite{fadili2025}. Specifically, tracking techniques, such as those leveraging Bayesian filters, provide predictions about future object orientation, position and size, thereby enhancing the effectiveness of the fusion process. NMS-based and distance-based association and fusion techniques generally do not perform any actual fusion in such scenarios. Instead, they simply retain the detected objects from both agents, resulting in a precision of approximately $50\%$. This essentially indicates that the number of false positives is equal to the number of true positives, highlighting the lack of an effective fusion process.

\section{Conclusion}
\label{sec:conclusion}
We introduced \textit{WLS-3D w/ CSBA-3D}, a late fusion framework for multi-agent 3D object association and fusion. By leveraging uncertainty-aware multi-score association and Weighted Least-Squares fusion, our method significantly improves translation, scale, and orientation metrics while maintaining perfect precision-recall performance. Extensive experiments confirm its superiority over state-of-the-art baselines, demonstrating robust performance under varying noise conditions, heterogeneous sensor configurations, and all object categories including Vulnerable Road Users (pedestrian, bicyle...). These results validate the effectiveness of our approach in real-world collaborative perception settings, establishing a new benchmark for late fusion in multi-agent collaborative perception systems.

We aim to extend our approach to accommodate asynchronous data fusion, addressing challenges where agents’ detections arrive with latency or at varying frame rates. Additionally, we plan to tackle misalignment issues between agents and sensors, classification inconsistencies where different agents assign conflicting object categories, as well as handling false positives and missed detections. Furthermore, we intend to broaden our evaluation by comparing our method with deep fusion approaches and validating the entire perception pipeline, including the detection stage (not covered in this study), on the latest cooperative perception datasets.

\bibliographystyle{IEEEtran}
\bibliography{main}

\end{document}